# On the existence and multiplicity of extensions in dialectical argumentation

**Bart Verheij**
Department of Metajuridica
Universiteit Maastricht[1]

**Abstract**

In the present paper, the existence and multiplicity problems of extensions are addressed. The focus is on extension of the stable type. The main result of the paper is an elegant characterization of the existence and multiplicity of extensions in terms of the notion of dialectical justification, a close cousin of the notion of admissibility. The characterization is given in the context of the particular logic for dialectical argumentation DEFLOG. The results are of direct relevance for several well-established models of defeasible reasoning (like default logic, logic programming and argumentation frameworks), since elsewhere dialectical argumentation has been shown to have close formal connections with these models.

## 1 INTRODUCTION

When a theory is interpreted in the context of logics for defeasible reasoning, e.g., in terms of the theory's extensions, it occurs for many such logics that there exist theories that cannot be interpreted at all or that have more than one interpretation.

For instance, in Reiter's (1980) well-known logic for default reasoning, a theory's extensions can be thought of as its interpretations. The simplest theory without extension consists of only the default $true : \neg p / p$. A basic example of a theory with more than one extension consists of the facts p and q and the defaults p: r / r and q : ¬r / ¬r.

For a good understanding of defeasible reasoning, it is natural to investigate under what circumstances theories are (so to speak) defeasibly interpretable, in the sense of having an extension, and under what circumstances their defeasible interpretation is ambiguous, in the sense of having more than one extension.

Several properties of theories have been discussed that guarantee the existence of extensions. For instance, in the context of his default logic, Reiter (1980) defined normal theories, that could be shown to have at least one extension. Etherington (1987) defined ordered theories and showed that ordered, semi-normal theories always have an extension (cf. also Papadimitriou and Sideri 1994). More results and references are for instance given by Gabbay *et al.* (1994) and Brewka *et al.*(1997).

In the present paper, the notorious extension existence and extension multiplicity problems are addressed in the context of a specific form of defeasible reasoning, viz. dialectical argumentation. In dialectical argumentation, statements are not only adduced as reasons for other statements, but also as reasons against. As a result of this possibility of simultaneously supporting and attacking statements, there can be statements that are justified (e.g. since there is a justifying reason for the statement) and statements that are defeated (e.g. since there is a defeating reason against the statement). Chesñevar *et al.* (2000) and Prakken and Vreeswijk (to appear) give overviews of models of dialectical argumentation. Formal models of dialectical argumentation are closely related to other nonmonotonic logics, as has especially been shown by Dung (1995) and Bondarenko *et al.* (1997). As a result, it comes as no surprise that the existence and multiplicity problems also arise in the context of dialectical argumentation. Dung (1995) and Bondarenko *et al.* (1997) give several relevant results in this respect (for different possible kinds of extensions).

In the present paper, the focus is on extensions of the stable type (cf. Gelfond and Lifschitz's (1988) stable models of logic programming, Dung's (1995) stable extensions of argumentation frameworks, and Bondarenko *et al.*'s (1997) stable extensions of assumption-based frameworks). The notion of dialectical justification is introduced and is shown to play a central role in the solution of the extension existence and multiplicity problems. The main results of the paper are characterizations of the existence of an extension and of the number of extensions in terms of dialectical justification. By a meta-analysis , it is shown that the notion of dialectical justification is vital in the characterization of the existence and the multiplicity of extensions.

---

[1] bart.verheij@metajur.unimaas.nl,
http://www.metajur.unimaas.nl/~bart/

The results are proven for a particular formalization of dialectical argumentation, viz. DEFLOG (Verheij 2000a, b). The system is related to my work on automated argument assistance (e.g., Verheij 1999). DEFLOG is formally closely related to Dung's (1995) and Bondarenko *et al.*'s (1997) formalizations. For Dung's (1995) argumentation frameworks, the connection is formally established in section 4 below. With respect to the logical language, DEFLOG differs from Dung's (1995) and Bondarenko *et al.*'s (1997) formalizations in that it allows the explicit expression in the logical object language of support and attack and of the defeat of a statement. As a result, support and attack become subject to dispute. The defeat of a statement is as a variant of negation, called *dialectical negation*. In contrast, Dung (1995) (who does not discuss support) uses a fixed, undisputable attack relation. Bondarenko *et al.* (1997) use a fixed, undisputable set of rules of inference to express support. They use a domain-dependent mapping of sentences to their contraries that already hints at the explicit expression of defeat in the logical object language.

By the already mentioned close formal relations between well-established models of defeasible reasoning (like default logic, logic programming and argumentation frameworks) and models of dialectical argumentation, the results proven for DEFLOG are of direct relevance for these models.

## 2 DIALECTICAL ARGUMENTATION

The context of the present work is dialectical argumentation. Here follows a brief introduction along the lines of my recent work on that topic (Verheij 1999, 2000a, b).

In dialectical argumentation, statements can not only support other statements, but also attack them. For instance, as a reason to support that Peter shot George, the statement can be made that some witness, say A, states that Peter shot George:

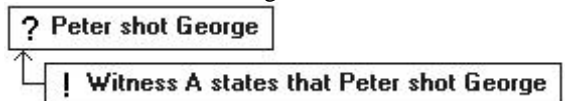

The exclamation mark indicates an assumed statement, the question mark a statement that is at issue. (The graphical presentation of arguments is based on previous work by the present author.) Here the issue that Peter shot George is settled (the statement is justified, as is indicated by the dark, bold font) since there is a justifying reason for it, namely A's testimony.

As a reason against the issue that Peter shot George, the statement can be made that some other witness, say B, states that the shooting did not take place:

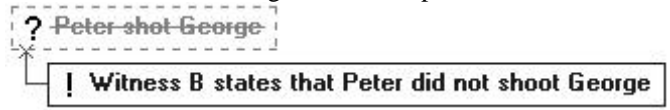

Assuming only B's testimony (not A's), the issue that Peter shot George is again settled, but this time the statement is defeated, as is indicated by the struck-through font. (Note that here B's testimony is used to argue *against* the statement that Peter shot George, and not to argue *for* the statement that Peter did *not* shoot George. As will be seen in the formal discussion below, arguing against a statement is treated as arguing for the statement's so-called dialectical negation, which is not necessarily equivalent to arguing for its ordinary negation.)

That some statement supports or attacks another statement can itself be at issue. For instance, it can be argued that A's testimony supports that Peter shot George since witness testimonies are often truthful:

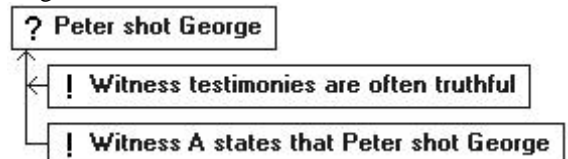

Likewise, a reason can be given to support that some statement *attacks* another statement.

A's unreliability can be adduced in order to attack that A's testimony supports that Peter shot George:

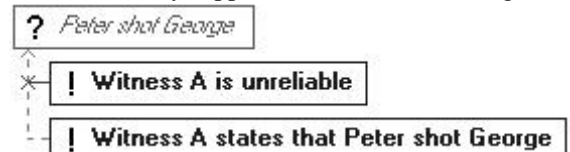

Here the issue that Peter shot George is unsettled, as is indicated by the light italic font, since it is not justified (e.g., by a justifying reason for it) nor defeated (e.g., by a defeating reason against it). Similarly, a reason can be given to attack that some statement attacks another statement.

## 3 DEFLOG - A LOGIC OF DIALECTICAL INTERPRETATION

### 3.1 THE DIALECTICAL INTERPRETATION OF THEORIES

The ideas on dialectical argumentation discussed in section 2 can be made formally precise in terms of the logical system DEFLOG (Verheij 2000a). Its starting point is a simple logical language with two connectives × and ~›. The first is a unary connective that is used to express the defeat of a statement, the latter is a binary connective that is used to express that one statement supports another. When φ and ψ are sentences, then ×φ (φ's so-called *dialectical negation*) expresses that the statement that φ is defeated, and (φ → ψ) that the statement that φ supports the statement that ψ. Attack, denoted as ⋈, is defined in terms of these two connectives: φ ⋈ ψ is defined as φ → ×ψ, and expresses that the statement that φ attacks the statement that ψ, or equivalently that the statement that φ supports that the statement that ψ is defeated. When p, q, r and s are elementary sentences, then p → (q → r), p → ×(q → ×r) and (p → q) → (p → ×(r → s)) are some examples of sentences. (For convenience, outer brackets are omitted.)

The central definition of DEFLOG is its notion of the *dialectical interpretation* of a theory. Formally, DEFLOG's dialectical interpretations of theories are a variant of Reiter's (1980) extensions of default theories, Gelfond and Lifschitz's (1988) stable models of logic programming, Dung's (1995) stable extensions of argumentation frameworks, and Bondarenko *et al.*'s (1997) stable extensions of assumption-based frameworks.[1]

A theory is any set of sentences, and when it is dialectically interpreted, all sentences in the theory are evaluated, either as justified or as defeated. (This is in contrast with the interpretation of theories in standard logic, where all sentences in an interpreted theory are assigned the same positive value, namely true, e.g., by giving a model of the theory.)

An assignment of the values justified or defeated to the sentences in a theory gives rise to a dialectical interpretation of the theory, when two properties obtain.

---

[1] In section 4, a formal connection with Dung's (1995) work is discussed. More relations between the formalisms mentioned are e.g. discussed by Dung (1995) and in the extended manuscript on which the present is based (Verheij 2000a). To guide intuition, the following may be useful. A default p : q / r (as in Reiter's 1980) would in DEFLOG be translated to two conditionals, viz. p → r and ¬q → ×(p →r). The second says that the former is defeated in case of ¬q. This corresponds to the intuition underlying the default that r follows from p as long as q can consistently be assumed. (Note however that the properties of ordinary negation ¬ are not part of DEFLOG proper.) A rule in logic programming p ← q, ~r corresponds in DEFLOG to two conditionals, viz. q → p and r → ×(q → p). The second says that q → p is defeated in case of r. This corresponds to the intuition underlying the program rule that p follows from q when r is not provable, but not when r is provable.

First, the justified part of the theory must be conflict-free. Second, the justified part of the theory must attack all sentences in the defeated part. Formally the definitions are as follows.

(i) Let T be a set of sentences and φ a sentence. Then T *supports* φ when φ is in T or follows from T by the repeated application of →-Modus ponens (i.e., from φ → ψ and φ, conclude ψ). T *attacks* φ when T supports ×φ.

(ii) Let T be a set of sentences. Then T is *conflict-free* when there is no sentence φ that is both supported and attacked by T.

(iii) Let Δ be a set of sentences, and let J and D be subsets of Δ that have no elements in common and that have Δ as their union. Then (J, D) *dialectically interprets* the theory Δ when J is conflict-free and attacks all sentences in D. The sentences in J are the *justified statements* of the theory Δ, the sentences in D the *defeated statements*.

(iv) Let Δ be a set of sentences and let (J, D) dialectically interpret the theory Δ. Then (Supp(J), Att(J)) is a *dialectical interpretation* or *extension* of the theory Δ. Here Supp(J) denotes the set of sentences supported by J, and Att(J) the set of sentences attacked by J. The sentences in Supp(J) are the *justified statements* of the dialectical interpretation, the sentences in Att(J) the *defeated statements*.

Note that when (J, D) dialectically interprets Δ and (Supp(J), Att(J)) is the corresponding dialectical interpretation, J is equal to Supp(J) ∩ Δ, and D to Att(J) ∩ Δ. It is convenient to say that a dialectical interpretation (Supp(J), Att(J)) of a theory Δ *is specified by* J.

The examples discussed in section 2 can be used to illustrate these definitions. Let s express Peter's shooting of George, a A's testimony, b B's testimony, t the truthfulness of testimonies, and u A's unreliability. Then the first example corresponds to the two-sentence theory {a, a → s}. The arrow in the figure corresponds to the sentence a → s. The theory has a unique extension in which all statements of the theory are justified. In the extension, one other statement is justified, viz. s. The second example corresponds to the theory {b, b → ×s}. The arrow ending in a cross in the figure corresponds to the sentence b → ×s. The theory has a unique extension in which again all sentences of the theory are justified. In the extension, there are two other interpreted statements, viz. ×s, which is justified, and s, which is defeated. (The reader may wish to check that the theory {b, b → ×s, s}, which is not conflict-free, has the same unique extension, but that one of the statements in the theory is defeated.) The third example corresponds to the theory {a, t, t → (a → s)}. In its unique extension, all statements of the theory are justified, and in addition a → s and s. The fourth example corresponds to the theory {a, u, u → ×(a → s)}. In its unique extension, a → s is defeated and s is not interpreted (i.e., neither justified nor defeated). (The theory {a, u, u → ×(a → s), a → s} is not conflict-free, but has the same unique extension.)

DEFLOG's connectives → and × are obviously reminiscent of propositional logic's connectives → and ¬. Also some of DEFLOG's definitions remind of propositional logic. These likenesses have been incorporated on purpose. In fact, DEFLOG has been carefully designed to be as close as possible to propositional logic (as the paradigmatic example of deductive logic), while retaining the essence of defeasible logic.[2]

DEFLOG's connectives are not equivalent to propositional logic's connectives. For instance, the set {p, ×p} is not inconsistent, in the dialectical sense: the theory {p, ×p} has a unique dialectical interpretation in which p is defeated and ×p justified. Of course {p, ¬p} is classically inconsistent. The theory {p, ×p} shows the essence of dialectical negation: the dialectical negation of a sentence in a sense 'prevails' over the sentence.[3] By this prevalence of dialectical negation, assumptions are only *prima facie* justified: a *prima facie* assumption is not actually justified when the dialectical negation of the assumption is (actually) justified.

The theory {p, ×p} also shows that dialectical interpretation is not simply maximal consistency: whereas the maximal consistent subset {×p} corresponds to a (the) dialectical interpretation, but {p} does not.

Verheij (2000a) gives much more information on DEFLOG, for instance, on different ways to adapt DEFLOG to incorporate the classical logical connectives. (DEFLOG's connectives → and × are not meant to *replace* the classical connectives; they express different concepts.)

It is not hard to see that DEFLOG is non-monotonic, for instance in the following sense: when a sentence is justified in some dialectical interpretation of a theory, it need not be in a dialectical interpretation of a larger theory. The simplest example is provided by the theories {p} and {p, ×p}. Both have only one dialectical interpretation. In the dialectical interpretation of {p}, p is justified, but in that of {p, ×p}, p is defeated (and ×p justified).

Notwithstanding the simple structure of DEFLOG's logical language (with only two connectives, viz. → and ×), many central notions of dialectical argumentation can be analyzed in terms of it. For instance, it is possible to define an inconclusive conditional (i.e., a conditional of which the consequent does not always follow when its antecedent obtains) in terms of DEFLOG's defeasible conditional (that is defeasible in the same way as any other statement). DEFLOG's expressiveness also allows an integrated analysis of Toulmin's (1958) warrants and backings and Pollock's (1987) undercutting and rebutting defeaters. A warrant and an undercutter can be seen as the support and attack, respectively, of the relation between a reason and its conclusion. Undercutting and rebutting defeaters become different instances of the general phenomenon of defeat. Cf. Verheij (2000a, 2001).

## 3.2 THEORIES WITHOUT EXTENSIONS AND THEORIES WITH SEVERAL EXTENSIONS

The examples of theories discussed above all had a unique extension. Several were examples of the following general property: a conflict-free theory always has a unique extension, namely the extension specified by the theory itself. The simplest theory that is not conflict-free with a unique extension is {p, ×p}. In its extension, p is defeated and ×p justified. Other important examples of theories that are not conflict-free, but do have a unique extension are {p, q, q → ×p} and {p, q, r, q → ×p, r → ×q}. In the former theory, the statement that p, is attacked by the statement that q. In its unique extension, q and ×p are justified and p is defeated. In the latter theory, a superset of the former, in addition to q's attack of p, r attacks q. In its unique extension, p, ×q and r are justified, and q is defeated. The theories together provide an example of *reinstatement*: a statement is first defeated, since it is attacked by a counterargument, but becomes justified by the addition of a counterattack, i.e., an attack against the counterargument. Here p is reinstated: it is first successfully attacked by q, but the attack is then countered by r attacking q.

There are however also theories with no or with several extensions:

(i) The three theories {p, p → ×p}, {p, p → q, ×q} and {$p_i$ | i is a natural number} ∪ {$p_j$ → ×$p_i$ | i and j are natural numbers, such that i < j} lack extensions.

---

[2] One result of this exercise is that I have come to believe that the concept of dialectical negation (along with the corresponding concept of dialectical interpretation) is the essential difference between a deductive and a defeasible logic.

[3] Note that the prevalence relation between a sentence p and its weak negation ~p in logic programming is exactly opposite to that between a sentence p and its dialectical negation ×p: in logic programming ~p can be assumed as long as p is not provable, while in dialectical argumentation p can be assumed as long as ×p is not justified.

For the latter theory, this can be seen as follows. Assume that there is an extension E in which for some natural number n $p_n$ is justified. Then all $p_m$ with m > n must be defeated in E, for if such a $p_m$ were justified, $p_n$ could not be justified. But that is impossible, for the defeat of a $p_m$ with m > n can only be the result of an attack by a justified $p_{m'}$ with m' > m. As a result, no $p_i$ can be justified in E. But then all $p_i$ must be defeated in E, which is impossible since the defeat of a $p_i$ can only be the result of an attack by a justified $p_j$ with j > i. (Note that any *finite* subset of the latter theory has an extension, while the whole theory does not. This shows a 'non-compactness' property[4] of extensions.)

(ii) The three theories $\{p, q, p \rightsquigarrow \times q, q \rightsquigarrow \times p\}$, $\{p_i, p_{i+1} \rightsquigarrow \times p_i \mid i$ is a natural number$\}$ and $\{\times^i p \mid i$ is a natural number$\}$ have two extensions. Here $\times^i p$ denotes, for any natural number i, the sentence composed of a length i sequence of the connective $\times$, followed by the constant p. (Note that each finite subset of the latter theory has a unique extension, showing another non-compactness property.)

Traditionally, the main example of multiple extensions is the Nixon diamond. In Reiter's (1980) default logic, it looks thus:

    q, r
    q : p / p
    r : ¬p / ¬p

These express that Nixon is a quaker and a republican, and that quakers are pacifists, while republicans are non-pacifists. Reiter's definitions give rise to two extensions. In one, p follows by the application of the first default, in the other, ¬p follows by the application of the first default.

It may be thought that its DEFLOG representation has the form $\{q, r, q \rightsquigarrow p, r \rightsquigarrow \times p\}$. However, this set does not have a dialectical interpretation. A good representation of the Nixon diamond consists of the following assumptions:

    q, r
    q $\rightsquigarrow$ p
    r $\rightsquigarrow$ ¬p
    (q & (q $\rightsquigarrow$ p)) $\rightsquigarrow$ ×(r $\rightsquigarrow$ ¬p)
    (r & (r $\rightsquigarrow$ ¬p)) $\rightsquigarrow$ ×(q $\rightsquigarrow$ p)

Here $(\varphi \& \psi) \rightsquigarrow \chi$ abbreviates $\varphi \rightsquigarrow (\psi \rightsquigarrow \chi)$. The connective ¬ is intended to express ordinary negation. The latter two conditionals express that when q justifies p (expressed by the conjunction of q and q $\rightsquigarrow$ p), it is defeated that r implies ¬p, and that when r is actually justified, it is defeated that q implies p. In each of the two dialectical interpretations of these assumptions, one of q $\rightsquigarrow$ p and r $\rightsquigarrow$ ¬p is defeated, the other justified. In the DEFLOG formalization, the conditionals q $\rightsquigarrow$ p and r $\rightsquigarrow$ ¬p stand for the 'application' of the corresponding Reiter defaults, while the conditionals (q & q $\rightsquigarrow$ p) $\rightsquigarrow$ ×(r $\rightsquigarrow$ ¬p) and (r & r $\rightsquigarrow$ ¬p) $\rightsquigarrow$ ×(q $\rightsquigarrow$ p) express when that application is blocked. The difference between the Reiter and the DEFLOG formalization has to do with the fact that Reiter's defaults are inconclusive (their consequent does not always follow when their antecedent obtains), while DEFLOG uses conditionals that are prima facie justified, just like other assumptions. Cf. also Verheij (2000a). Note that the opposition between p and ¬p is not represented in terms of dialectical negation (i.e., as p and ×p): ordinary negation and dialectical negation are different notions.

## 3.3 DIALECTICALLY JUSTIFYING ARGUMENTS

Before we proceed to the notion of dialectical justification, some terminology needs to be introduced.

(i) A set of sentences is an *argument* when it is conflict-free. If Δ is a set of sentences, a Δ-*argument* is an argument that is a subset of Δ.

(ii) Let φ be a sentence. An argument C is an *argument for* φ if C supports φ. An argument C is an *argument against* φ if C attacks φ. The sentences in an argument C are also called its *premises*, the sentences φ such that C supports φ, its *conclusions*.

(iii) An argument C *attacks* an argument C' if C attacks a sentence in C'.

(iv) Arguments C and C' are *compatible* when C ∪ C' is an argument, and otherwise *incompatible*. The arguments in a collection $\{C_i\}_{i \in I}$ are *compatible* if their union $\cup_{i \in I} C_i$ is an argument, otherwise *incompatible*.

In the following figure, three arguments are graphically suggested.

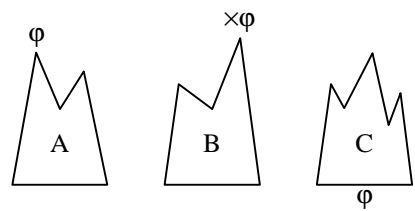

The bottoms of the alpine shapes consist of the premises of the argument; the tops are the conclusions. Argument A has conclusion φ, argument B conclusion ×φ and argument C has premise φ. B attacks C, but not necessarily A (since φ might not be a premise of A). A and B are incompatible, and B and C are too.

---

[4] A property P of sets is called compact if a set S has property P whenever all its finite subsets have the property. Cf. the compactness of satisfiability in first-order predicate logic.

When a theory has a dialectical interpretation, the set of sentences of the theory that are justified in the interpretation, clearly form an argument. It has a special property:

*Proposition*
> Let E be an extension of a theory Δ. Then J(E) ∩ Δ is a Δ-argument that attacks any Δ-argument C that is incompatible with J(E) ∩ Δ. Here J(E) denotes the set of justified statements of the extension E.

*Proof:* Since E is an extension, J(E) ∩ Δ is conflict-free. Hence a Δ-argument C that is incompatible with J(E) ∩ Δ cannot be a subset of J(E) ∩ Δ since J(E) ∩ Δ is not incompatible with any of its subsets. Therefore there is a sentence φ in C that is not in J(E) ∩ Δ. Since E is an extension, it is in D(E), the set of defeated sentences of the extension E. But for any sentence φ in D(E) it holds by the definition of extensions that J(E) ∩ Δ attacks φ, and therefore attacks C. QED

Arguments with the property that J(E) ∩ Δ has in the proposition above, are said to be dialectically justifying:

(v) A Δ-argument C is *dialectically justifying* with respect to Δ if and only if C attacks every Δ-argument C' that is incompatible with C.

(vi) A sentence φ is *dialectically justifiable* with respect to a set of sentences Δ if and only if there is a Δ-argument C for φ that is dialectically justifying with respect to Δ. Such an argument C is then called a *dialectical justification of* φ, and C *dialectically justifies* φ with respect to Δ. A sentence φ is *dialectically defeasible* with respect to Δ if and only if ×φ is dialectically justifiable with respect to Δ. If C is a dialectical justification of φ, then the argument C *dialectically defeats* φ with respect to Δ.

(vii) A sentence φ is *dialectically interpretable* with respect to a set of sentences Δ if and only if it is dialectically justifiable or dialectically defeasible with respect to Δ. A sentence φ is *dialectically ambiguous* with respect to a set of sentences Δ if and only if it is both dialectically justifiable and dialectically defeasible with respect to Δ.

The argument {p, r, r → ×q} dialectically justifies p with respect to the theory {p, q, r, q → ×p, r → ×q}. The argument {p} does not dialectically justify p since the incompatible argument {q, q → ×p} is not attacked. The argument {r, r → ×q} dialectically defeats q with respect to the theory.

The sentences p and q are dialectically ambiguous with respect to the theory {p, q, p → ×q, q → ×p} since the argument {p, p → ×q} dialectically justifies p and dialectically defeats q, and likewise for q.

The sentence p is not dialectically interpretable with respect to the theory {p, p → ×p}.

Note that when an argument is dialectically justifying with respect to a theory, it dialectically justifies all the sentences it supports.

## 3.4 THE EXISTENCE AND MULTIPLICITY OF EXTENSIONS

When a theory has a dialectical interpretation, all sentences in the theory are dialectically interpretable. In other words, dialectical justification is a kind of 'local' dialectical interpretation. This is an immediate corollary of the proposition proven in section 3.3:

*Corollary*
> Let E be an extension of the theory Δ. Then all sentences in the theory are dialectically justifiable or dialectically defeasible with respect to Δ.

*Proof:* By the proposition of section 3.3, J(E) ∩ Δ dialectically justifies or defeats all sentences in Δ. QED

Note that the corollary gives a necessary condition for the existence of an extension: when there is a sentence in a theory that is not dialectically interpretable, there cannot be an extension. The corollary can explain all examples of theories without extensions that have been encountered above: in all, there is a sentence that is not dialectically interpretable. Nevertheless the condition in the corollary is *not* sufficient for the existence of an extension, as the theory Δ = {p, q, p → ×q, q → ×p, r → ×r, s, s → ×s, p → ×r, q → ×s} shows. It has no extension. Nevertheless all sentences in the theory are dialectically justifiable or defeasible with respect to Δ. The Δ-argument {p, p → ×q, p → ×r} dialectically justifies p and dialectically defeats q and r, while {q, q → ×p, q → ×s} dialectically justifies q and dialectically defeats p and r.

The notion of dialectical justification plays the central role in the main theorem of the present paper, that shows exactly under which circumstances a theory has an extension. One additional definition is needed.

(viii) Let C be an argument. A sentence φ is *dialectically justifiable in the context* C with respect to a theory Δ if it is supported by a dialectically justifying argument of the theory that contains C, and *dialectically defeasible in the context* C if ×φ is supported by a dialectically justifying argument that contains C.

Using this terminology, the main theorem can be formulated:

**Theorem**
>A theory $\Delta$ has an extension if and only if there is an argument C in the context of which all sentences in $\Delta$ are either dialectically justifiable or dialectically defeasible with respect to the theory, but not both.

(The proof follows below.) In other words, a theory has an extension if and only if there is a context in which all sentences of the theory are dialectically interpretable, while none is dialectically ambiguous. The theorem is closely related to the corollary above that says that the dialectical interpretability of all sentences of a theory is necessary for the existence of an extension. The theorem says that the dialectical interpretability of all sentences *in a context with no dialectical ambiguities* is both necessary and sufficient for the existence of an extension. After fixing all choices allowed by a dialectically ambiguous sentence in the theory, it suffices for the existence of an extension that all sentences in the theory are either dialectically justifiable or dialectically defeasible. The example that showed why the dialectical interpretability of all sentences of a theory is not sufficient for the existence of an extension, shows what can go wrong: the dialectical justification of one sentence (or its dialectical negation) need not be compatible with that of another when there is a dialectical ambiguity. In other words, the dialectical justification of sentences can depend on the particular choice allowed by a dialectical ambiguity. Dialectical justifications that require different choices cannot be 'glued' to form an extension.

Three properties of dialectical justification are essential in the proof of the theorem:

*Proposition*
(i) *Localization:* Let E be an extension of a theory $\Delta$. Then there is a collection $\{C_i\}_{i \in I}$ of arguments that covers $J(E) \cap \Delta$ (i.e., $J(E) \cap \Delta$ is equal to $\cup_{i \in I} C_i$), that are dialectically justifying with respect to the theory.
(ii) *Union:* If C and C' are compatible arguments, that are dialectically justifying with respect to a theory $\Delta$, then also $C \cup C'$ is dialectically justifying with respect to the theory. (Similarly, for collections of dialectically justifying arguments: the union of a compatible collection of dialectically justifying arguments is again dialectically justifying.)
(iii) *Separation at the base:*[5] If C and C' are incompatible arguments, that are dialectically justifying with respect to a theory $\Delta$, then there is a sentence in $\Delta$ that is both dialectically justifiable and defeasible with respect to $\Delta$. (Similarly, for collections of dialectically justifying arguments: given an incompatible collection of dialectically justifying arguments, there is a sentence in the theory that is both dialectically justifiable and defeasible.)

*Proof:* Localization follows from the proposition at the beginning of this section: it shows that $J(E) \cap \Delta$ is itself dialectically justifying with respect to $\Delta$. The union property (for pairs of arguments) is seen as follows. Let C and C' be compatible dialectically justifying arguments, and let the argument C'' be incompatible with $C \cup C'$. Assume first that C'' is incompatible with C. Then clearly C attacks C''. Assume second that C'' is compatible with C. Then C' is incompatible with the argument $C \cup C''$, and therefore attacks it. Since C and C' are compatible, it then follows that C' attacks C''. The proof of the general case of the union property requires some extra care, but is similar. The property of separation at the base follows directly from the definition of dialectical justification: when C and C' are dialectically justifying and incompatible, they attack each other. Then there is a sentence in each (and therefore in the theory itself) that is attacked by the other. The general case of the separation property can be reduced to the case of pairs of arguments. QED

*Proof of the main theorem:* First let E be an extension of $\Delta$. Then by the localization property $J(E) \cap \Delta$ can be covered by arguments that are dialectically justifying with respect to $\Delta$. By the union property, it then follows that $J(E) \cap \Delta$ is also dialectically justifying. (In fact, the proof of the corollary at the beginning of the section directly shows that $J(E) \cap \Delta$ is dialectically justifying.) As a result, $J(E) \cap \Delta$ is a context as in the theorem since by the fact that $J(E) \cap \Delta$ is dialectically justifying and by the definition of extensions all sentences in $\Delta$ are dialectically interpretable in the context of $J(E) \cap \Delta$, and since by the fact that $J(E) \cap \Delta$ is conflict-free there is no dialectically ambiguous sentence in that context. Second let C be a context as in the theorem, and let, for all sentences $\varphi$, $C_\varphi$ be a $\Delta$-argument dialectically justifying or defeating $\varphi$ in the context C. The collection of the $C_\varphi$ is compatible since by the property of separation at the base there would otherwise be a sentence in the theory that is dialectically ambiguous in the context C. By the union property, the union of the $C_\varphi$ is dialectically justifying. It specifies an extension of $\Delta$. QED

The proof shows that extensions can be built by 'gluing' dialectically justifying arguments. This suggests that a (set-theoretically minimal) argument that dialectically justifies a sentence, is a kind of 'dialectical proof' of the sentence. Similarly, such a dialectical proof of the dialectical negation of a sentence is a kind of 'dialectical refutation' of the sentence.

---
[5] The property is called separation *at the base* since the dialectically ambiguous sentence can be found in the theory itself.

The following theorem provides a general answer to the extension existence and multiplicity problems in the context of dialectical argumentation. It is a corollary of the main theorem above:

**Theorem**
Let *n* be a natural (or cardinal) number (possibly 0). A theory Δ has exactly *n* extensions if and only if *n* is equal to the maximal number of mutually incompatible arguments C in the context of which all sentences in Δ are either dialectically justifiable or dialectically defeasible with respect to the theory, but not both.

## 4 DUNG'S ARGUMENTATION FRAMEWORKS AND ADMISSIBILITY

Dung's (1995) argumentation frameworks are a fruitful abstraction of ideas from nonmonotonic reasoning and logic programming. Here it is shown how Dung's argumentation frameworks can be mimicked in DEFLOG. Since Dung has shown that his argumentation frameworks have close formal connections with well-established models of defeasible reasoning, such as Reiter's (1980) default logic and logic programming, the results on DefLog presented here become of direct relevance for these models. Moreover it is shown why Dung's notion of admissibility cannot in general replace that of dialectical justification in the characterizations of the existence and of the number of extensions of a theory proven above.

Formally, an argumentation framework consists of a set, its elements called *arguments*, and a binary relation on that set, the *attack* relation. When (A, B) is in the attack relation, the argument A is said to attack B.

In Dung's work, the notion of admissibility is central. It is closely related to DEFLOG's dialectical justification. Using DEFLOG's terminology, an argument C is *admissible* with respect to a theory Δ if C attacks any Δ-argument attacking it. This definition of admissibility depends of course on DEFLOG's particular notions of argument and attack. There is however a straightforward way of mimicking Dung's argumentation frameworks in DEFLOG for which this definition of admissibility is indeed an extrapolation of Dung's admissibility, as follows.

Let each argument of an argumentation framework be an elementary sentence in DEFLOG's language. Then an argumentation framework can be translated to a theory in DEFLOG by taking the union of the set of arguments in the framework and the set of sentences of the form A ⇁ ×B, for any element (A, B) of the attack relation of the framework. In addition, it is easy to restrict DEFLOG's language in such a way that any theory in this restricted language corresponds to an argumentation framework in Dung's sense: simply allow only elementary sentences and sentences of the form φ ⇁ ×ψ, where φ and ψ are elementary. Let's call sentences in this restricted sense *Dung sentences* and theories consisting of Dung sentences *Dung theories*.

It is now straightforward to check that several of Dung's notions coincide with DEFLOG's under this translation. Some care is needed however since certain terms have different meanings in Dung's work and in DEFLOG. For instance, the use of the term 'argument' is different. Conflict-free sets of arguments (in Dung's sense) correspond however with conflict-free sets of Dung sentences (in DEFLOG's sense), Dung's admissible sets of arguments correspond to the admissible arguments of Dung theories (in DEFLOG's sense), and Dung's stable extensions of argumentation frameworks correspond with DEFLOG's extensions of Dung theories. In the extended manuscript on which the present paper is based (Verheij 2000a), these results are formally established. The proofs are straightforward.

For theories using DEFLOG's full language, dialectical justification and admissibility are easily seen to be different notions, but on the restricted language of Dung's frameworks, the notions coincide:

*Proposition*
Let Δ be a Dung theory. Then a Δ-argument is dialectically justifying with respect to Δ if and only if it is admissible with respect to Δ.

*Proof:* Dialectically justifying arguments are always admissible. (This does not depend on Δ being a Dung theory.) Let now C be an admissible argument, and let C' be an argument incompatible with C. Since C and C' consist of Dung sentences, the incompatibility of C and C' implies that C attacks C' or that C' attacks C. In case C' attacks C, also C attacks C' since C is admissible. This shows that C is dialectically justifying. QED

Note that by this result the theorems on the extension existence and multiplicity problems can *for Dung theories* be rephrased in terms of admissibility instead of dialectical justification. This is not the case for theories in general. Then the notion of dialectical justification is essential. The key point is that admissibility does not have all of the properties used in the proof of the main theorem of this paper. These properties are localization, union and separation at the base.

The analogues of these properties for admissibility can be found by replacing 'dialectically justifying' by 'admissible' in the formulation of the properties. For instance, the union property (for pairs of arguments) for admissibility reads thus: if C and C' are compatible arguments, that are admissible with respect to a theory Δ, then also C ∪ C' is admissible with respect to the theory. Separation at the base becomes (again for pairs of arguments): if C and C' are incompatible arguments, that are admissible with respect to a theory Δ, then there are opposites φ and ψ in the theory, such that C supports φ and C' supports ψ.

It is not hard to see that admissibility has the localization and union properties, but lacks the property of separation at the base.

For instance, that for admissibility, the property of separation at the base does not obtain, can be seen by inspection of the theory $\{p_1, p_1 \rightarrow q, p_2, p_2 \rightarrow (q \rightarrow \times q)\}$. With respect to the theory, there are four admissible arguments with a maximal number of elements, viz. each three-element subset of the theory. (Note that each argument of the theory is admissible since there are no attacking arguments.) Any pair of these arguments is incompatible, yet there is no sentence that is defeated by an argument, let alone by an admissible argument, as is required by the property of separation at the base.

That the localization property obtains for admissibility is straightforward: since, when E is an extension of a theory Δ, J(E) ∩ Δ is dialectically justifying with respect to Δ, J(E) ∩ Δ is certainly admissible.

The proof of the union property for admissibility is almost trivial since any attack of the union of a collection of arguments is also an attack of one of the arguments in the collection.

Inspection of the proof of the main theorem of the paper shows that the property of separation at the base is only used in the 'if'-part. The 'only if'-part indeed has an analogue for admissibility since it only uses localization and union. The theory $\{p_1, p_1 \rightarrow q, p_2, p_2 \rightarrow (q \rightarrow \times q)\}$ (the counterexample against the property of separation at the base) shows that the analogue of the 'if'-part is in fact not true. All sentences in the theory are 'admissibly justifiable', i.e., supported by an admissible argument, since any argument of the theory is admissible. No sentence in the theory is 'admissibly defeasible', i.e., attacked by an admissible argument, since there is no attacking argument at all. Still, the theory has no extension.

Verheij (2000a) expands this meta-analysis for other results (e.g., concerning so-called dialectically preferred and admissibly preferred arguments, i.e., those dialectically justifying or admissible arguments that are maximal with respect to set inclusion) and for other notions that are similar to dialectical justification.

Bondarenko *et al.* (1997) have used admissibility in their discussion of an abstract, argumentation-theoretic approach to default reasoning. Their setting is just as Dung's (1995) related to DEFLOG's, yet they focus on deductive systems. Interestingly, whereas in DEFLOG dialectical negation × is treated as an ordinary connective, Bondarenko *et al.* consider the question which sentences are the contraries of others as part of the domain theory (as the mapping from sentences to their contraries is explicitly represented in their assumption-based frameworks). It seems that the notion of dialectical justification can be directly transplanted to their system. For the reasons, discussed here and in section 3.4, it can be expected that dialectical justification has better properties for analyzing assumption-based frameworks than admissibility.

## 5 CONCLUSION

A characterization of the existence of extensions (in the context of dialectical argumentation) has been established that shows that the dialectical interpretability of all sentences in a theory suffices for the existence of an extension provided that all dialectical ambiguities in the theory are fixed. A characterization of the multiplicity of extensions immediately follows.

Whether these characterizations should be regarded as solutions to the extension existence and multiplicity problems depends on one's taste. The notion of dialectical interpretation (a typical specimen of the genus of stable extensions) has been connected to the notion of dialectical justification, itself not a very simple notion. The complexity of the problems does however not suggest a really simple solution.

In an important sense, the connection can however be regarded as a reduction: dialectical interpretation is a *global* notion (of interpreting a theory as a whole), whereas dialectical justification is a *local* notion (of interpreting a sentence with respect to a theory). The characterization shows how the global and local notions are connected. The characterization makes it formally precise that local interpretability is a requirement of global interpretability, and that global ambiguity has its roots in local ambiguity.